\documentclass[letterpaper, 10 pt, conference]{ieeeconf}
\IEEEoverridecommandlockouts            
\let\labelindent\relax                  

\pdfoutput=1
\usepackage[utf8]{inputenc}
\usepackage{graphicx} 
\usepackage{amsmath,amsfonts,amssymb,booktabs,siunitx,leftidx} 
\usepackage[hidelinks]{hyperref} 
\usepackage[nolist]{acronym}  
\usepackage[inline]{enumitem} 
\usepackage{flushend} 
\usepackage{pgfplots} 
\pgfplotsset{compat=1.8}
\usepackage{multirow}
\usepackage{array}
\usepackage{cite}
\usepackage{subcaption}
\usepackage{silence}
\newcommand{\figref}[1]{\hyperref[#1]{Fig.~\ref*{#1}}}
\newcommand{\tabref}[1]{\hyperref[#1]{Tab.~\ref*{#1}}}
\newcommand{\secref}[1]{\hyperref[#1]{Sec.~\ref*{#1}}}

\DeclareMathOperator*{\argmin}{argmin}

\def\gt{ground truth}
\def\ie{\textit{i.e.},}
\def\eg{\textit{e.g.},}
\def\Eg{\textit{E.g.},}
\def\etal{\textit{et al.}}

\newacro{vlm}[VLM]{visual-language model}
\newacro{llm}[LLM]{large language model}
\newacro{3dsg}[3DSG]{3D scene graph}

\title{\LARGE \bf REACT: Real-time Efficient Attribute Clustering and Transfer for Updatable 3D Scene Graph}

\author{Phuoc~Nguyen, Francesco~Verdoja, Ville~Kyrki%
	\thanks{This work was supported by the Research Council of Finland (decision 354909). P. Nguyen, F. Verdoja, and V. Kyrki are with School of Electrical Engineering, Aalto University, Espoo, Finland. {\tt\small \{firstname.lastname\}@aalto.fi}}}
\begin{document}
\makeatletter
\let\@oldmaketitle\@maketitle
\renewcommand{\@maketitle}{\@oldmaketitle
	\setcounter{figure}{0}
	\vspace{1em}
	\centering
	\includegraphics[width=\linewidth]{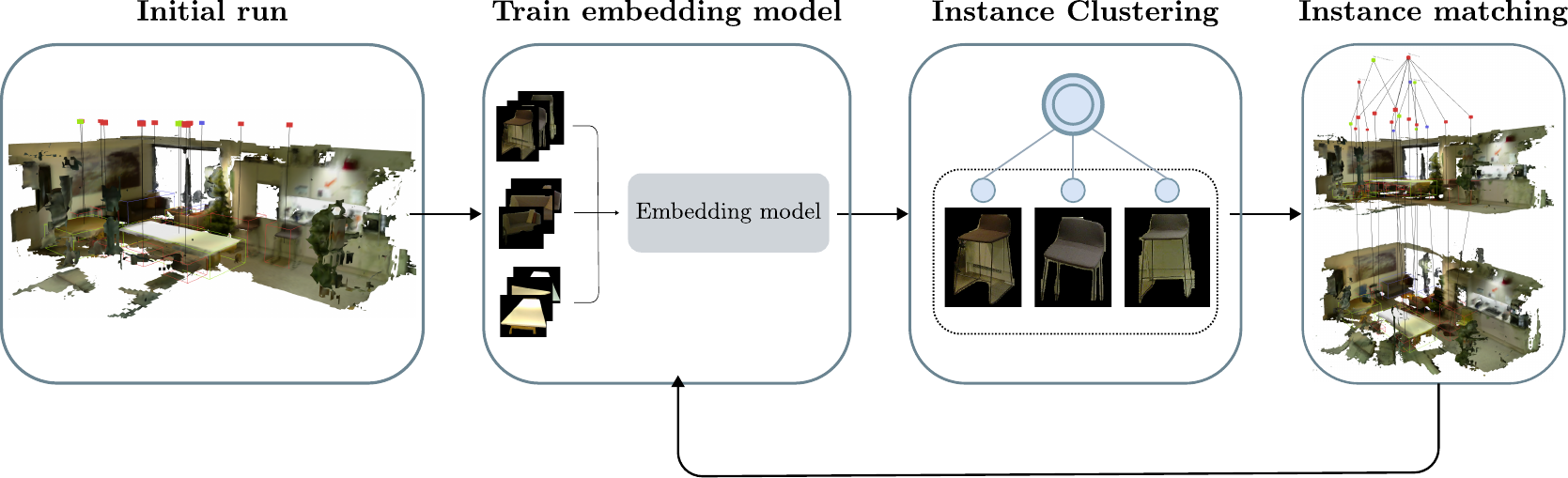}
	\captionof{figure}{\label{fig:cover}We propose REACT, a method to efficiently update object nodes in a~\acl{3dsg} in real-time by leveraging computationally efficient visual features. In this work, we pursue two objectives. Firstly, we aim to cluster identical object instances to facilitate attribute sharing among multiple object nodes within a~\acl{3dsg}. Secondly, we tackle the challenge of instance matching in long-term, semi-static scenarios.}}
\makeatother

\maketitle

\begin{abstract}
	Modern-day autonomous robots need high-level map representations to perform sophisticated tasks. Recently, \acp{3dsg} have emerged as a promising alternative to traditional grid maps, blending efficient memory use and rich feature representation. However, most efforts to apply them have been limited to static worlds. This work introduces REACT, a framework that efficiently performs real-time attribute clustering and transfer to relocalize object nodes in a~\ac{3dsg}. REACT employs a novel method for comparing object instances using an embedding model trained on triplet loss, facilitating instance clustering and matching. Experimental results demonstrate that REACT is able to relocalize objects while maintaining computational efficiency. The REACT framework's source code will be available as an open-source project, promoting further advancements in reusable and updatable~\acp{3dsg}\footnote{\url{https://github.com/aalto-intelligent-robotics/REACT.git}}.
\end{abstract}
\acresetall

\section{Introduction}

Nowadays, autonomous mobile robots are expected not only to navigate between locations but also to understand the environments they traverse. In recent years, the research community has developed various advanced map representations as alternatives to traditional occupancy grid maps. These innovations aim to enable robots' scene comprehension by incorporating semantic spatial information. Among these developments,~\acp{3dsg} have emerged as a promising solution, offering efficient memory storage while maintaining a rich scene representation. ~\cite{armeni3DSceneGraph2019a}.

Significant efforts have been devoted to enriching~\acp{3dsg} by incorporating more complex information into object nodes, such as embeddings from~\acp{vlm}~\cite{guConceptGraphsOpenVocabulary3D2023,werbyHierarchicalOpenVocabulary3D2024}. These rich features facilitate the execution of higher-level tasks, allowing mobile robots to respond to complex commands more effectively. However, computing the features comes with substantial computational expense, with some methods requiring days of processing or multiple GPUs to complete~\cite{werbyHierarchicalOpenVocabulary3D2024,guConceptGraphsOpenVocabulary3D2023}. Additionally, many constructed maps lack updatability and reusability, restricting their applicability to simulators or static environments. The lack of efficient methods to relocalize objects on the fly has the drawback that, if the environment has changed, the rich features embedded in the nodes of the previously built~\acp{3dsg} might be unusable when a robot revisits an environment.

In this work, we propose a method to leverage computationally efficient features for real-time instance matching and relocalization. By transferring pre-computed attributes and features, robots can efficiently adapt to new environments while reusing the rich information in the constructed~\acp{3dsg}. Moreover, in real-world settings, there are often many identical objects, such as sets of tables and chairs in study halls or similar mugs in kitchens. We propose that embracing these visual similarities during~\ac{3dsg} construction can potentially reduce storage and computational overhead. This improvement can be achieved by sharing attributes between identical nodes such as images, 3D representations (meshes, point clouds), and neural embeddings. An illustration of the proposed method is shown in \figref{fig:cover}.

In particular, we propose the REACT framework, the core contributions of which include:
\begin{itemize}
	\item A novel approach for recognizing similar objects and distinguishing different ones using an embedding model trained on triplet loss.
	\item A method for clustering object instances with similar appearances to facilitate attribute sharing.
	\item A framework for matching and relocalizing object nodes in a~\ac{3dsg} while detecting missing or novel objects in real-time as the robot revisits an environment.
	\item Experimental evidence that clustering object instances can enhance instance matching performance, demonstrated by comparing our proposed method with a greedy approach that does not employ attribute clustering.
	\item The REACT framework's source code will be made publicly available as an open-source project, promoting further advancements in reusable and adaptable~\acp{3dsg}.
\end{itemize}

\section{Related works}

\subsection{3D scene graphs}

\acp{3dsg} have recently emerged as an efficient yet expressive representation for both indoor and outdoor environments~\cite{hughesHydraRealtimeSpatial2022a,hughesFoundationsSpatialPerception2024,rosinol3DDynamicScene2020a,greveCollaborativeDynamic3D2024}. By abstracting objects and spatial concepts such as rooms and buildings into nodes, and connecting these nodes with edges that represent their relationships,~\acp{3dsg} effectively scale to larger scenes~\cite{hughesHydraRealtimeSpatial2022a,hughesFoundationsSpatialPerception2024}. Notably, Hughes \etal{}~\cite{hughesHydraRealtimeSpatial2022a} have demonstrated the feasibility of constructing such rich metric-semantic graphs in real-time, enhancing the potential for their application in robotic tasks.

Efforts have been made to leverage~\acp{3dsg} as enablers for executing natural language commands. Rana \etal{} presented Sayplan~\cite{ranaSayPlanGroundingLarge2023}, which converts pre-built~\acp{3dsg} into JSON files, using them as input for a~\ac{llm} to perform semantic searches on sub-graphs, exploiting the hierarchical nature of~\ac{3dsg}. More recently, ConceptGraphs~\cite{guConceptGraphsOpenVocabulary3D2023} and HOV-SG~\cite{werbyHierarchicalOpenVocabulary3D2024} have pioneered embedding object nodes with features from visual-linguistic models (VLMs). The authors demonstrated the ability to query for task-related objects using the capabilities of LLMs, applying these to various downstream tasks.

However, a major limitation in all of these approaches is the lack of reusability of the constructed scene graphs. Many works operate under the assumption of a static environment when querying information from the scene graph~\cite{ranaSayPlanGroundingLarge2023}, or use interim measures like querying potential locations using an LLM~\cite{guConceptGraphsOpenVocabulary3D2023}. In this work, we propose REACT as a potential step forward in enhancing the reusability and updatability of~\acp{3dsg}.

\subsection{Object change detection}

In computer vision, several recent works have addressed object matching and relocalization. Wald \etal{}~\cite{waldRIO3DObject2019b} propose a multi-scale triplet network to detect instance-level changes in a scene using TSDF patches extracted from RGB-D data. The model identifies feature keypoints on the source object and the entire target scene, then computes an optimal transformation to relocalize the source object. More recently, Zhu \etal{}~\cite{zhuLivingScenesMultiobject2024} propose a framework to parse a 3D indoor environment as an evolving scene. Their approach reconstructs objects with increasing accuracy as more temporal scans become available, and compares their embeddings with their respective counterparts in other scenes. These methods generally require preprocessed point clouds and are not optimized for real-time deployment in robotic systems.

To detect and track object changes, Bore \etal{}~\cite{boreDetectionTrackingGeneral2019} assume that semi-static objects primarily move short distances within a location, with a probability of ``jumping'' to another location. Their work samples the posterior for objects' global movements and tracks local changes analytically using Kalman filters. By forming probabilistic models for objects, their work can handle input ambiguity such as visually similar objects. In contrast, our work leverages visual similarity to enhance matching performance while still optimizing the likelihood of the new positions of objects.

Another line of research that explores measures to help robot adapt to evolving environments is to detect evidence of changes within the built maps and update them accordingly. Panoptic Multi-TSDF~\cite{schmidPanopticMultiTSDFsFlexible2022} maintains a collection of object-wise multi-scale TSDFs, tracking object changes by counting the number of inconsistent voxels over overlapping sub-maps. POCD~\cite{qianPOCDProbabilisticObjectLevel2022a} updates a volumetric map by introducing an object association module to find correlations between observations and tracked objects, estimating geometric changes in the scene. Most recently, Schmid \etal{} introduces Khronos~\cite{schmidKhronosUnifiedApproach2024b}, which aims to build a spatio-temporal representation of a scene to infer the states of objects at different time points. While these frameworks mark objects not seen in their previous locations as absent or removed, our work actively matches each object instance with its correspondence as the robot revisits an area.

\section{Problem Formulation}
\label{sec:problem_formulation}

The problem formulation is based on the following assumptions:
\begin{enumerate}
	\item A robot performs multiple mapping sessions within the same indoor environment, with no dynamic agents present during any session.
	\item A localization method, such as 2D SLAM~\cite{kohlbrecherFlexibleScalableSLAM2011,macenskiSLAMToolboxSLAM2021}, visual-inertial odometry~\cite{huaiRobocentricVisualInertial2022} or \gt{} information (for simulators), provides the robot's pose during the mapping process.
	\item  Scene changes are limited to rigid transformations. As defined in~\cite{waldRIO3DObject2019b}, rigid changes encompass the addition, removal, and relocation of objects within the scene between two mapping sessions. Detection of non-rigid changes, \eg{} changes in objects' appearances, lighting conditions, or operational states is beyond the scope of this work.
	\item The environment may contain multiple identical objects (\eg{} identical chairs in a meeting room, or a plate set on a kitchen table).
\end{enumerate}

Formally, at time $t$, the scene consists of a collection of $m$ object instances $\mathcal{O}_t=\{I_t^1,\dots,I_t^m\}$, where $I$ denotes a single object instance. The scene evolves over time such that the set of $q$ objects at time $t+1$ is $\mathcal{O}_{t+1}=\{{I}^{1}_{t+1},\dots,{I}^{q}_{t+1}\}$.

Let us define an identity function $\mathcal{I}(I^i_t,I^j_{t+1}) \in \{0,1\}$, such that $\mathcal{I}(I^i_t,I^j_{t+1}) = 1$ when $I^i_t$ re-appears at $t+1$ as $I^j_{t+1}$, and $\mathcal{I}(I^i_t,I^j_{t+1}) = 0$ otherwise. In other words, $I^i_t$ and $I^j_{t+1}$ represent the exact same object recorded at different time. Then, the relationship between times $t$ and $t+1$ is defined in terms of sets $M_{t:t+1}$, $A_{t:t+1}$, and $N_{t:t+1}$. The \textit{Matched} set $M_{t:t+1}$ includes object instances that remain in the scene, but might have moved; formally $M_{t:t+1} = \{(I_t^i,I_{t+1}^j) \mid I_t^i \in \mathcal{O}_{t} \wedge I_{t+1}^j \in \mathcal{O}_{t+1} \wedge \mathcal{I}(I_t^i,I_{t+1}^j) = 1\}$. The \textit{Absent} set $A_{t:t+1} \subseteq \mathcal{O}_{t}$ includes those that were in $\mathcal{O}_t$ but do not appear at time $t+1$; and the \textit{New} set $N_{t:t+1} \subseteq \mathcal{O}_{t+1}$ those that only appear at time $t+1$.

The objective is to identify the set of instance associations $\{M_{t:t+1},A_{t:t+1},N_{t:t+1}\}$ between the object sets $\mathcal{O}_t$ and $\mathcal{O}_{t+1}$ and update our map such that all instances in $M_{t:t+1}$ have their position updated, all instances in $A_{t:t+1}$ are removed, and all instances in $N_{t:t+1}$ are added.

\section{Methodology}

To address the problem outlined in \secref{sec:problem_formulation}, we propose REACT, whose pipeline is illustrated in~\figref{fig:cover}. Initially, a robot patrols a scene and constructs an initial~\ac{3dsg} (we will discuss the process of building the~\ac{3dsg} in \secref{subsec:3dsg_building}). Then, we manually group images of identical instances and train an embedding model to facilitate comparisons between object nodes. Subsequently, we utilize this trained model to cluster the object nodes. When the robot returns to the same environment, as it constructs a new~\ac{3dsg} of the environment, it clusters the newly created nodes and matches them with their correspondences from the previous~\ac{3dsg}. Finally, it updates the object nodes association.

\subsection{Identity function}
\label{sec:identity}

In practical scenarios involving multiple identical objects, it is impossible to evaluate the identity function $\mathcal{I}$ as discussed in \secref{sec:problem_formulation} unless the environment is under constant observation. However, it is a logical assumption that semi-static objects undergo minimal movement between consecutive mapping sessions. For example, while it is impossible to correctly match indistinguishable chairs in a kitchen, it is reasonable to expect that a chair placed next to a particular table would be more likely to be the one found around that same table at the next session than another one further away. Therefore, we propose to approximate $\mathcal{I}$ by decomposing it into two sub-problems: visual similarity and motion minimization.

Firstly, we assess if two objects exhibit strong visual similarity; \ie{} $V(I^i, I^j) \leq \gamma$, where $V(\cdot) \geq 0$ quantifies the visual difference between two object instances and $\gamma$ is a predetermined threshold parameter. Evaluating visual similarity while mapping is challenging due to different viewpoints and occlusion across sessions. Furthermore, as previously discussed, we assume scenes contain multiple identical object instances that, barring their positions, share numerous attributes such as visual appearance, semantics, geometry, and functionality. To harness this similarity and improve matching robustness, we organize identical object instances into instance clusters, denoted as $\mathcal{C}$, treating each object as part of a collective entity. This clustering approach enables us to gather more comprehensive information about indistinguishable objects by aggregating multiple instances of the same type. Thus, across sessions, we approximate visual similarity as: $V(I^i_t, I^j_{t+1}) \approx \tilde{V}(\mathcal{C}^k_t, \mathcal{C}^l_{t+1}),\ I^i_t \in \mathcal{C}^k_t,\ I^j_{t+1} \in \mathcal{C}^l_{t+1}$. We will discuss how to build these clusters in the next subsection.

Secondly, after matching clusters $\mathcal{C}^k$ and $\mathcal{C}^l$ in terms of visual similarity, we associate instances within these clusters by minimizing the total distance traveled by all matched object instances, formulated as an linear sum assignment problem~\cite{crouseImplementing2DRectangular2016}. Let us define the assignment function $\mathbf{1}_{kl}: \mathcal{C}^k \times \mathcal{C}^l \to \{0,1\}$ mapping instances from $\mathcal{C}^k$ to $\mathcal{C}^l$.
Assuming $|\mathcal{C}^k| \geq |\mathcal{C}^l|$, minimizing the total distance traveled is solved by:
\begin{align}
    \begin{split}
        \mathbf{1}^*_{kl} = \argmin_{\mathbf{1}_{kl}} {\sum_{I^i}^{\mathcal{C}^k}\sum_{I^j}^{\mathcal{C}^l}{D\left(p^i, p^j\right)\mathbf{1}_{kl}(I^i,I^j)}}\\
        \text{subject to:} \sum_{I^i \in \mathcal{C}^k} \mathbf{1}_{kl}(I^i,I^j) = 1\;\forall I^j \in \mathcal{C}^l\\
        \sum_{I^j \in \mathcal{C}^l} \mathbf{1}_{kl}(I^i,I^j) \leq 1\;\forall I^i \in \mathcal{C}^k
    \end{split}
    \label{fun:optim_matching}
\end{align}
where $p^i$ represents the position of instance $I^i$, $p^j$ denotes the position of instance $I^j$, and $D \geq 0$ is a spatial distance function.

After this minimization, we obtain the approximation of the identify verification, \ie{} $\mathcal{I}(I^i,I^j) \approx \mathbf{1}^*_{kl}(I^i,I^j)$ with $I^i \in \mathcal{C}^k$ and $I^j \in \mathcal{C}^l$.

\subsection{Instance matching}

During each mapping session at time $t$, the robot constructs a set of $m$ object instances $\mathcal{O}_t$. Based on visual similarity, we form $n$ instance clusters, such that $\mathcal{O}_t=\mathcal{C}^1_t \cup \cdots \cup \mathcal{C}^n_t, n \leq m$. As similar instances are merged into an instance cluster $\mathcal{C}^i_t, i \in [1, n]$, they share attributes, including the semantic class $c_i$, the collection of raw images of the object instance taken at different viewpoints $\mathcal{V}_i=\{v_1,\dots,v_o\}$, as well as the average visual embedding $f_i$ that describe them (elaborated in \secref{subsec:instance_matching}). To enable instance matching, each instance cluster keeps track of the position history of each object instance, formally labeled $\mathcal{P}^j=\left\{p_t^j \mid t \in [0,\dots,T], j \in [0,\dots,m] \right\}$ for instance $j$.

In a subsequent mapping session, a new set of instance clusters $\mathcal{O}_{t+1}=\{\mathcal{C}^1_{t+1},\dots,\mathcal{C}^q_{t+1}\}$ is generated. After visually comparing and matching the instance clusters from both sets, we infers the movements of objects between time $t$ and $t+1$ by optimizing \eqref{fun:optim_matching}.

In our implementation, the robot operates in compact indoor environments, and the observed object-level changes are typically minor relocalizations. As a result, the Euclidean distance is a natural choice for $D(\cdot)$ in \eqref{fun:optim_matching}. However, the selection of the distance metric $D(\cdot)$ may vary depending on the specific application and operating environment. \Eg{} for operations encompassing a larger area with multiple rooms and floors, one may build a traversability graph and let $D(\cdot)$ be the shortest travel distance between two points using the A* algorithm.

\section{Implementation}
\subsection{3D Scene Graph with Instance Memory and Object Clusters}
\label{subsec:3dsg_building}

Our~\ac{3dsg} updating system builds upon the foundation of Hydra~\cite{hughesFoundationsSpatialPerception2024,hughesHydraRealtimeSpatial2022a}. As the robot patrols an environment, it incrementally constructs a hierarchical~\ac{3dsg} in real-time. The nodes of this graph represent various spatial concepts across different layers. These layers, from top to bottom, include buildings, rooms, traversable spaces, and objects. The edges of the graph denote inclusion relationships among the nodes.

Our work focuses on the targets the robot will interact with during its assignments, specifically the object nodes. Since our method operates at the instance level, we modified Hydra to segment the object mesh using results from an off-the-shelf instance segmentation model.

Moreover, we extended the~\ac{3dsg} from Hydra in three ways. Firstly, drawing inspiration from~\cite{changGOATGOAny2023}, each object node maintains its own Instance Memory, which includes raw images of each object instance taken from multiple viewpoints. Each instance view produces a binary mask associated with its corresponding RGB scene image. Secondly, to facilitate instance matching, each node's Instance Memory contains identifiable visual embeddings, which will be discussed in \secref{subsec:instance_matching}. Finally, when multiple nodes of identical objects form an instance cluster, a ``cluster'' node is added and their instance memories are combined, resulting in an extensive image library with more object viewpoints and an average of all instances' visual embeddings.

\subsection{Learning for Attribute Clustering and Transfer}
\label{subsec:instance_matching}

Clustering object instances and re-identifying them in subsequent scans necessitate a method for measuring the visual similarity between objects, \ie{} $V(\cdot)$ as described in \secref{sec:identity}. To this end, we implement an end-to-end structure as described in FaceNet\cite{schroffFaceNetUnifiedEmbedding2015}, employing its pipeline. The embedding model's backbone is based on EfficientNet\cite{tanEfficientNetRethinkingModel2019}, resulting in a visual embedding $f(\cdot)$ for each instance view. After the first mapping session, we train this embedding model employing a triplet network architecture and minimizing the triplet loss
\begin{align}
	L = \sum_{i=1}^{N}{\left[\lVert f_i^a - f_i^p \rVert_2^2 - \lVert f_i^a - f_i^n \rVert_2^2 + \alpha \right]}
\end{align}
where $f^a$ is the anchor patch embedding, $f^p$ is the positive patch embedding, and $f^n$ is the negative patch embedding. The margin $\alpha$ indicates the amount of separation between anchor-positive pairs and anchor-negative pairs. In this work, we set $\alpha=1$.

Intuitively, minimizing the triplet loss reduces the Euclidean distance between visual embeddings of similar objects, regardless of viewing angles, while maximizing the distance between embeddings of different objects. To optimize training efficiency, we utilize the online triplet mining strategy popularized by~\cite{schroffFaceNetUnifiedEmbedding2015}, which aims to generate semi-hard negative samples for each anchor-positive pair and train on them.

As demonstrated in \figref{fig:cover}, the dataset for training the model is formed after the robot's first visit to an environment. Using the framework described in \secref{subsec:3dsg_building}, we construct a~\ac{3dsg} with object nodes containing raw RGB images of objects from different viewpoints. As our methodology focuses exclusively on the appearance of objects; thus, we mask out background elements in all comparison images. We interactively group objects with similar appearances into instance clusters, create training and validation datasets, and train the model with them. Empirically, we train the model for 30 epochs using Adam as the optimizer, with an initial learning rate of 0.001. Since the robot may observe some objects only briefly during its first scan, we augment the training dataset by horizontal flipping followed by random rotations between $-10^{\circ}$ and $10^{\circ}$ with uniform probability.

The trained embedding model then generates visual embeddings for each object node by retrieving the median of all RGB images in its Instance Memory. By thresholding the Euclidean distance between their respective embeddings, the object nodes are clustered and their embeddings are averaged into the cluster node. When matching instances during subsequent visits to the same environment at time $t+1$, new object nodes are also clustered and compared with their respective counterparts from time $t$. As each cluster $k$ contains an averaged embedding $f^k$ from all views of each instance, we select the Euclidean distance between two embeddings as the visual difference measurement $V(\cdot)$ (see \secref{sec:problem_formulation})
\begin{align}
    V\left(\mathcal{C}^k, \mathcal{C}^l\right) = \lVert f^k - f^l \rVert_2^2
\end{align}

This approach offers two main benefits. Firstly, it substantially reduces the computational complexity, particularly in environments with multiple similar objects. Secondly, the comparisons are more robust. Specifically, instead of comparing an object to a previous instance that might have been observed from a distance or with limited viewing angles, we compare it to an averaged visual embedding that encapsulates a more comprehensive representation of the object.

\subsection{Offline and online instance matching}

\begin{figure*}
	\includegraphics[width=\textwidth]{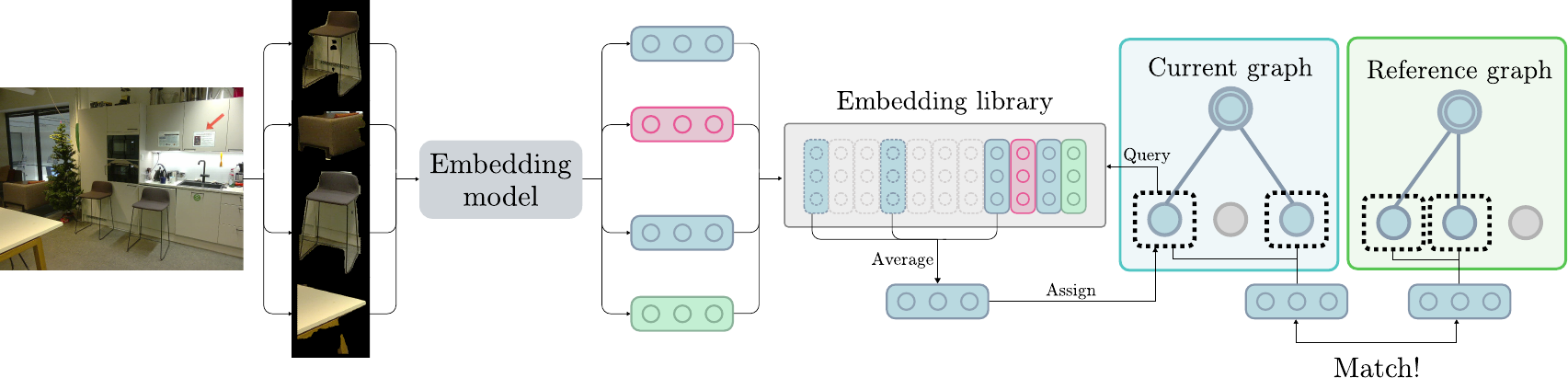}
	\caption{Illustration of the online matching process. As the robot gathers sensor data, the embedding model generates visual embeddings and stores them in an embedding library $E$. Each node of the current~\ac{3dsg} may query for the average embedding from $E$. After the nodes on the current graph are clustered, they are compared against clusters of the same semantic class in the reference graph and nodes from the reference graph are updated.}
	\label{fig:online}
\end{figure*}

As outlined in \secref{sec:problem_formulation}, we assume that objects experience minimal movement between mapping sessions. To this end, for each cluster in $t+1$ that matches in $t$, we perform instance matching using a modified Jonker-Volgenant algorithm with no initialization~\cite{jonkerShortestAugmentingPath1988,crouseImplementing2DRectangular2016} implemented in Scipy to optimize \eqref{fun:optim_matching}.

When conducting instance matching offline, the process is straightforward, as all information about the objects is available. During online operations, the robot incrementally gathers information about the environment, necessitating an adaptive process. \figref{fig:online} illustrates REACT's matching process in online mode. Given that the embedding model is the most computationally intensive component of our framework, we maintain a queryable embedding library $E$ to store embeddings for all views of objects encountered during the session. As object nodes are formed, they query their respective visual embeddings from $E$ and retrieve the median value when comparing. It is important to note that due to the incremental nature of the process, matching may require several registered views of an object to be accurate. Consequently, each time the~\ac{3dsg} updates, the entire process---including attribute clustering and comparing embeddings---is re-executed on all nodes with the newly gathered information. Thus, given the same amount of registered input, the results will eventually converge to the same as the offline version. Since these processes are CPU-based, they should not significantly compromise the performance of the pipeline as we will demonstrate in \secref{subsubsec:runtime}.

\section{Evaluation}
\subsection{Experimental set-up}

To evaluate our proposed method, we aim to address the following research questions:

\begin{itemize}
	\item How effectively does the learned embedding model recognize similar objects and differentiate between distinct ones?
	\item Does attribute clustering improve the system's performance and accuracy?
	\item Can the entire pipeline accurately match instances in real-time as the robot patrols the environment?
\end{itemize}

We evaluate our framework, REACT, using one simulated and three real-world datasets. The simulated flat dataset, introduced by Schmid \etal{}~\cite{schmidPanopticMultiTSDFsFlexible2022}, consists of two trajectories within a flat where objects are moved, added, and removed between runs, providing \gt{} segmentation, depth maps, camera poses, and change information for benchmarking. For~\ac{3dsg} construction, we focus on classes undergoing changes, excluding cups and plates due to their small size and distant visibility. Although the 3RScan data set consisting of various scenes with rigid object changes exists~\cite{waldRIO3DObject2019b}, at the time of writing, we could not use it due to an open problem with that software preventing us from obtaining the \gt{} instance segmentation information for the RGB images.

For our real-world datasets, data is collected using a Hello Stretch 2 mobile platform equipped with an Astra 2 RGB-D camera. Data collection occurs in three locations within our department building: the coffee room and two study halls. All scenes are mapped twice, with objects moved, removed, or added between sessions. For all experiments except for the runtime evaluation, we perform instance matching offline. The details of all our datasets are shown in \tabref{tab:scenes}. The ground truth data for these datasets are recorded based on the~\acp{3dsg} built by our method described in \secref{subsec:3dsg_building}. Thus, we do not account for errors from incorrect~\ac{3dsg} construction, \eg{} incorrect semantic labels, and splitting or merging of object nodes.

\begin{table*}
	\centering
	\caption{\label{tab:scenes}Description of the scenes used in our experiments. The numbers in brackets represent the category based on visual appearance, \eg{} 2 chairs(1,2) means 2 chairs of type 1 and 2 chairs of type 2}
	\small
	\begin{tabular}{lccc}
		\toprule
		\textbf{Scene}                       & \textbf{Objects}                                      & \textbf{New}                             & \textbf{Absent}                         \\
		\midrule
		\multirow{2}{*}{\textit{Flat}}       & 2 chairs(1), 1 table(1,2), 2 pictures(1,2), 1 bed(1), & 1 chair(2), 1 journal(3), 1 bed(2),      & \multirow{2}{*}{1 journal(2), 1 bed(1)} \\
		                                     & 1 picture(3,4,5), 1 lamp, 1 journal(1,2)              & 1 table(3), 1 laptop, 1 coffee machine   &                                         \\
		                                     &                                                       &                                          &                                         \\
		\textit{LabFront}                    & 15 chairs(1), 3 tables(1)                             & 1 chair(2,3,4)                           & 3 chairs(1)                             \\
		                                     &                                                       &                                          &                                         \\
		\multirow{2}{*}{\textit{CoffeeRoom}} & 7 chairs(1), 1 chair(2), 2 chairs(3)                  & \multirow{2}{*}{1 chair(4), 1 chair(5)}  & \multirow{2}{*}{1 chair(1), 1 chair(3)} \\
		                                     & 1 table(1,2,3), 2 couches                             &                                          &                                         \\
		                                     &                                                       &                                          &                                         \\
		\multirow{2}{*}{\textit{StudyHall}}  & 25 chairs(1), 3 chairs(2), 2 chairs(3),               & \multirow{2}{*}{2 chairs(1), 1 chair(4)} & 1 chair(2), 2 chairs(3),                \\
		                                     & 10 tables(1), 1 table(2), 2 couches                   &                                          & 1 table(1), 1 couch                     \\
		\bottomrule
	\end{tabular}

\end{table*}

We utilize YOLO11~\cite{yolo11_ultralytics} for object semantics in real environments, without retraining the instance segmentation model. From the 80 pre-trained COCO classes~\cite{linMicrosoftCOCOCommon2014}, we select couch, dining table, and chair. Experiments were conducted on a laptop with an Intel I7-11800H CPU and an NVIDIA GeForce RTX 3070 Mobile GPU. Stretch 2's 2D LiDAR, combined with the SLAM toolbox~\cite{macenskiSLAMToolboxSLAM2021}, provides the robot's localization.

\subsection{Experimental results and ablation study}
\label{subsec:exp}

\begin{table}
	\centering
	\caption{\label{tab:vsgreedy}Change detection accuracy on multiple scenes with optimal visual difference threshold.}
	\small
	\resizebox{\columnwidth}{!}{\begin{tabular}{llcccc}
			\toprule
			\textbf{Scene}                       & \textbf{Method}         & $\mathbf{F1}^{M}\uparrow$ & $\mathbf{F1}^{N}\uparrow$ & $\mathbf{F1}^{A}\uparrow$ & $\mathbf{\sum{D}}(m)\downarrow$ \\
			\midrule
			\multirow{2}{*}{\textit{Flat}}       & \textbf{REACT}          & 1.0                       & 1.0                       & 1.0                       & 6.19                            \\
			                                     & \textbf{w/o clustering} & 1.0                       & 1.0                       & 1.0                       & 6.19                            \\
			\midrule
			\multirow{2}{*}{\textit{LabFront}}   & \textbf{REACT}          & 1.0                       & 1.0                       & 1.0                       & 4.76                            \\
			                                     & \textbf{w/o clustering} & 1.0                       & 1.0                       & 1.0                       & 4.76                            \\
			\midrule
			\multirow{2}{*}{\textit{CoffeeRoom}} & \textbf{REACT}          & \textbf{1.0}              & 1.0                       & \textbf{1.0}              & \textbf{4.58}                   \\
			                                     & \textbf{w/o clustering} & 0.96                      & 1.0                       & 0.67                      & 6.96                            \\
			\midrule
			\multirow{2}{*}{\textit{StudyHall}}  & \textbf{REACT}          & \textbf{1.0}              & \textbf{1.0}              & \textbf{1.0}              & \textbf{26.56}                  \\
			                                     & \textbf{w/o clustering} & 0.98                      & 0.90                      & 0.86                      & 45.26                           \\
			\bottomrule
\end{tabular}}
\end{table}

\subsubsection{Change detection accuracy}

We evaluate the change detection performance of REACT against a baseline version without attribute clustering. This alternative version utilizes a greedy strategy by iteratively matching object nodes with their highest visual correspondences. Both configurations use the same embedding model with an optimal visual difference threshold $\gamma$. Given the nature of models trained on triplet loss~\cite{schroffFaceNetUnifiedEmbedding2015}, the values of $\gamma$ for each embedding model need to be selected individually. In our evaluations, we explore $\gamma$ ranging from 0 to 5 in 0.2 increments and identify the optimal thresholds for each configuration to maximize the aggregated F1 score for all sets: Matched, Asbsent, and New. \tabref{tab:vsgreedy} presents two metrics: the F1-score of the Matched ($F1^M$), New ($F1^A$), and Absent  ($F1^N$) sets, along with the total travel distance of objects between scenes in metres $\left(\sum D \right)$, as it relates to optimizing our matching function in \eqref{fun:optim_matching}. In this experiment, matching is considered successful when an object matches with an identical one.

\begin{figure}
	\centering
	\includegraphics[width=0.48\textwidth]{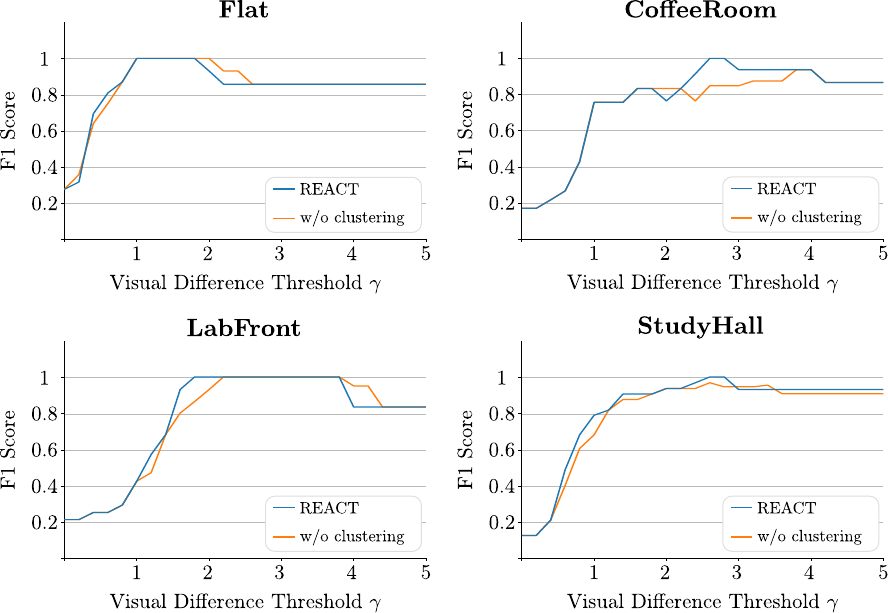}
	\caption{F1 scores over thresholds of REACT and its non-clustering baseline on all scenes.}
	\label{fig:f1_over_thresh}
\end{figure}

\begin{figure}
	\centering
	\includegraphics[width=0.45\textwidth]{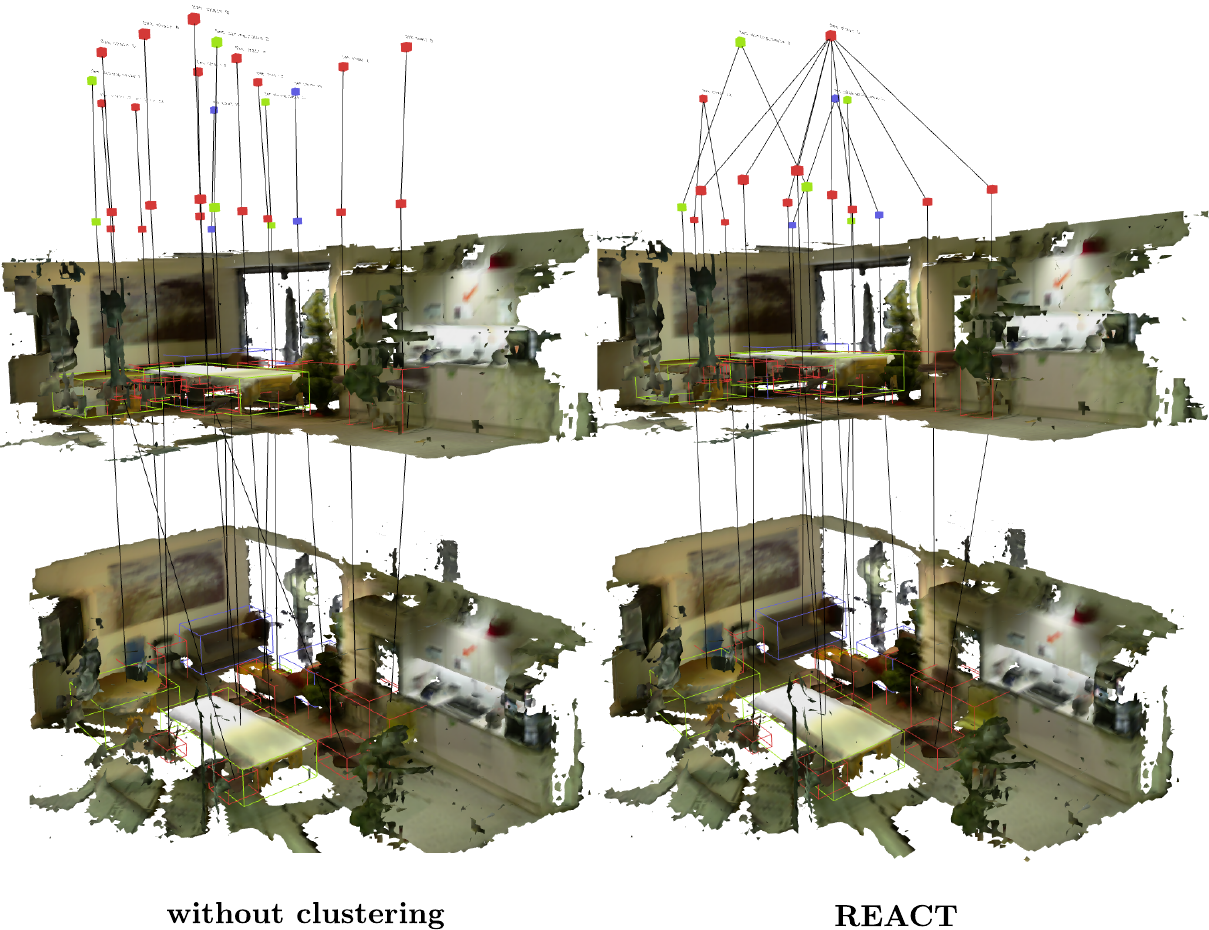}
	\caption{Qualitative comparison of change detection on the scene \textit{CoffeeRoom}.}
	\label{fig:qualitative_cmp}
\end{figure}

A comparison of the performance of both configurations across all thresholds is illustrated in \figref{fig:f1_over_thresh}. The results indicate that REACT maintains consistent performance across most thresholds. With optimal threshold selection, REACT matches or exceeds the performance of the non-clustering configuration across all available scenes. REACT's advantages become more apparent on large scenes with a high density of identical objects in varying orientations like \textit{StudyHall}. \figref{fig:qualitative_cmp} provides a qualitative comparison of the two methods on the scene \textit{CoffeeRoom}. The optimal threshold for REACT was found to be $\gamma=2.8$, while the non-clustering configuration had a threshold of $\gamma=3.8$. In the non-clustering configuration, the greedy matching may lead to suboptimal matches, as demonstrated by the matching of two chairs from opposite ends of the table despite the existence of better alternatives. A higher value for $\gamma$ in this configuration facilitates the matching of more identical chairs, but it also increases the likelihood of false positive matches, such as the gray chair being matched with the yellow one. Conversely, REACT correctly matches all identical instances. Moreover, by optimizing objects' travel distances post-matching, the results are more accurate and sensible, minimizing unnecessary significant shifts in the objects' positions.

\subsubsection{Ablation Study: Triplet Model Matching Performance}

\begin{table}
	\centering
	\caption{\label{tab:superglue}Instance matching performance.}
	\small
	\begin{tabular}{lcccc}
		\toprule
		Method                      & Precision     & Recall        & F1            & Avg. FPS       \\
		\midrule
		\textbf{SuperGlue}          & 0.82          & 0.45          & 0.59          & 4.45           \\
		\textbf{SuperGlue (masked)} & 0.80          & 0.64          & 0.71          & 4.45           \\
		\textbf{Ours}               & \textbf{0.97} & \textbf{0.98} & \textbf{0.98} & \textbf{86.99} \\
		\bottomrule
	\end{tabular}
\end{table}

To gauge the embedding model's ability to recognize similar objects between scans, we compare it against SuperGlue~\cite{sarlinSuperGlueLearningFeature2020}, a learning-based feature-matching method. This comparison employs the pre-trained indoor model with default configurations and a matching threshold of 6.0, as similarly applied in~\cite{changGOATGOAny2023}. We conduct two separate experiments for SuperGlue, one with the background masked out---the same setup as our method---and one with background context as proposed in the original work.

For this analysis, we use a~\ac{3dsg} built from the validation scans across all datasets and perform node matching analogous to the attribute clustering process. SuperGlue matches nodes when a single pair of instance views is matched. As shown in \tabref{tab:superglue}, our method significantly outperforms SuperGlue in matching accuracy and speed. Notably, while SuperGlue effectively detects similar objects, it struggles to differentiate distinct ones. Moreover, although Superglue performs better when provided with the contextual background of objects~\cite{changGOATGOAny2023}, our approach involves comparing objects with their identical counterparts in varying locations, resulting in different background contexts. Consequently, this leads to suboptimal performance when compared to utilizing SuperGlue on images with the background removed.

\begin{figure}
	\centering
	\includegraphics[width=0.48\textwidth]{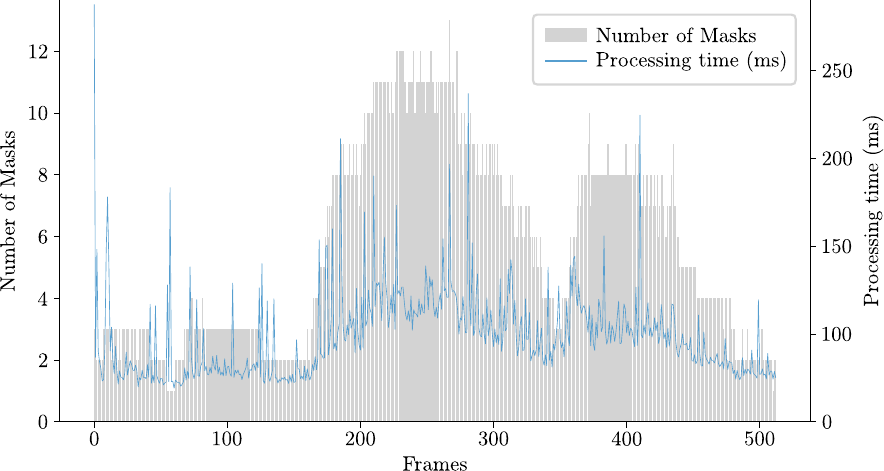}
	\caption{Runtime of embedding model relative to the number of instance masks per frame}
	\label{fig:emb_perf}
\end{figure}

\subsubsection{Online runtime}
\label{subsubsec:runtime}

As our method builds on the foundations of Hydra~\cite{hughesHydraRealtimeSpatial2022a}, which can incrementally construct a~\ac{3dsg} in real-time, we will evaluate the performance of the embedding model and instance matching in online matching mode. Using EfficientNet-B2 as the backbone with an input image size of $224 \times 224$, the embedding time was roughly 11ms per instance mask \cite{tanEfficientNetRethinkingModel2019}. \figref{fig:emb_perf} shows the embedding model's online matching performance on a re-scan in the \textit{CoffeeRoom} scene, peaking at about 200ms for 13 masks. For object node processing, which includes visual embedding comparisons and position history optimization, the CPU-based operations result in negligible latency. While there is still room for optimization, REACT achieves an average frame rate of 3 to 4Hz when mapping with a reconstruction resolution of 0.025m. It is worth noting that the input stream of RGB-D images for~\ac{3dsg} construction operates at 5Hz, and the~\ac{3dsg} construction by Hydra averages at 4Hz.

\section{Discussions}

As demonstrated in our experiments in \secref{subsec:exp}, REACT can reuse a~\ac{3dsg} and transfer attributes in real-time. The most notable transferable features in our work are the library of images and the accumulated average visual embeddings. By combining these features with attribute clustering, our method can enhance an object's information pool with memories of its past states and data from similar objects.

However, the list of shareable and transferable features extends beyond just images and visual embeddings. A practical extension is the inclusion of more powerful, albeit computationally expensive, features, \eg{}~\ac{vlm} embeddings. These could be efficiently transferred to new locations, enabling the robot to match the user queries while executing tasks in a dynamic environment. The works of Gu \etal{}~\cite{guConceptGraphsOpenVocabulary3D2023} and Werby et al~.\cite{werbyHierarchicalOpenVocabulary3D2024} have demonstrated the efficacy of such features, and we believe our work provides a link to bridge the usability gaps in these~\ac{3dsg} frameworks.

REACT’s shared attributes also benefit object reconstruction. As the robot traverses the environment, it collects point clouds of objects, which might be incomplete due to occlusions or sensor noise. By aggregating nodes of similar objects, it becomes possible to reconstruct more accurate representations. Improved object reconstruction, in turn, enhances the construction and maintenance of~\ac{3dsg} by offering more precise node boundaries and reducing noise.

While REACT represents a promising step towards adaptability for mobile robots in changing environments, it still has some limitations. A common failure mode for REACT stems from errors in scene graph construction, such as splitting a single object into multiple object nodes, merging nearby objects into one, or failing to register some objects. Accurately constructing object nodes in a scene graph is particularly challenging in environments with many adjacent objects, especially when there is noisy instance segmentation and unreliable sensor depth maps. Potential areas for improvement include, but are not limited to, utilizing more accurate segmentation models, applying multi-resolution object representation to allow for a more fine-grained representation of smaller objects and avoid faulty object merging~\cite{schmidPanopticMultiTSDFsFlexible2022}, and sensor fusion techniques to achieve more accurate point clouds.

Additionally, REACT's performance depends on the visual difference threshold, which requires some amount of parameter tuning. Moreover, our method necessitates pre-training the embedding model during the robot's first visit to a scene, which impacts its ability to generalize. Exploring the potential for generalization in visual-based instance matching using larger datasets and more advanced model architectures could be an interesting avenue for future research.
\section{Conclusion}

In this work, we present REACT, a novel framework to cluster object instances and leverage this clustering to update object nodes in a~\ac{3dsg} in real-time as a robot revisits a scene. Additionally, we propose an efficient image-based method for comparing objects' appearances. We evaluated our method through experiments conducted in a simulated environment and various real-world settings using data gathered from a real robot platform. Our experimental results provide evidence that clustering object instances and averaging their visual embeddings can enhance instance matching performance over a non-clustering approach while maintaining real-time processing speed. Our work constitutes a first step towards reusable and adaptable spatio-semantic representations, which are a necessary component for robots to operate in rich dynamic human environments.

\section{Acknowledgement}

The authors would like to acknowledge the use of Grammarly's generative AI tool for improving readability across all sections, as well as Aalto AI Assistant, which uses GPT 4o, for brainstorming the title of this work.
\bibliographystyle{IEEEtran}
\bibliography{IROS25_PNguyen}

\begin{thebibliography}{10}
\providecommand{\url}[1]{#1}
\csname url@samestyle\endcsname
\providecommand{\newblock}{\relax}
\providecommand{\bibinfo}[2]{#2}
\providecommand{\BIBentrySTDinterwordspacing}{\spaceskip=0pt\relax}
\providecommand{\BIBentryALTinterwordstretchfactor}{4}
\providecommand{\BIBentryALTinterwordspacing}{\spaceskip=\fontdimen2\font plus
\BIBentryALTinterwordstretchfactor\fontdimen3\font minus \fontdimen4\font\relax}
\providecommand{\BIBforeignlanguage}[2]{{%
\expandafter\ifx\csname l@#1\endcsname\relax
\typeout{** WARNING: IEEEtran.bst: No hyphenation pattern has been}%
\typeout{** loaded for the language `#1'. Using the pattern for}%
\typeout{** the default language instead.}%
\else
\language=\csname l@#1\endcsname
\fi
#2}}
\providecommand{\BIBdecl}{\relax}
\BIBdecl

\bibitem{armeni3DSceneGraph2019a}
I.~Armeni, Z.-Y. He, J.~Gwak, A.~R. Zamir, M.~Fischer, J.~Malik, and S.~Savarese, ``{{3D Scene Graph}}: {{A Structure}} for {{Unified Semantics}}, {{3D Space}}, and {{Camera}},'' in \emph{Proceedings of the {{IEEE}}/{{CVF International Conference}} on {{Computer Vision}}}, 2019, pp. 5664--5673.

\bibitem{guConceptGraphsOpenVocabulary3D2023}
Q.~Gu, A.~Kuwajerwala, S.~Morin, K.~M. Jatavallabhula, B.~Sen, A.~Agarwal, C.~Rivera, W.~Paul, K.~Ellis, R.~Chellappa, C.~Gan, C.~M. {de Melo}, J.~B. Tenenbaum, A.~Torralba, F.~Shkurti, and L.~Paull, ``{{ConceptGraphs}}: {{Open-Vocabulary 3D Scene Graphs}} for {{Perception}} and {{Planning}},'' Sep. 2023.

\bibitem{werbyHierarchicalOpenVocabulary3D2024}
A.~Werby, C.~Huang, M.~B{\"u}chner, A.~Valada, and W.~Burgard, ``Hierarchical {{Open-Vocabulary 3D Scene Graphs}} for {{Language-Grounded Robot Navigation}},'' Jun. 2024.

\bibitem{hughesHydraRealtimeSpatial2022a}
N.~Hughes, Y.~Chang, and L.~Carlone, ``Hydra: {{A Real-time Spatial Perception System}} for {{3D Scene Graph Construction}} and {{Optimization}},'' Jun. 2022.

\bibitem{hughesFoundationsSpatialPerception2024}
N.~Hughes, Y.~Chang, S.~Hu, R.~Talak, R.~Abdulhai, J.~Strader, and L.~Carlone, ``Foundations of spatial perception for robotics: {{Hierarchical}} representations and real-time systems,'' \emph{The International Journal of Robotics Research}, p. 02783649241229725, Feb. 2024.

\bibitem{rosinol3DDynamicScene2020a}
A.~Rosinol, A.~Gupta, M.~Abate, J.~Shi, and L.~Carlone, ``{{3D Dynamic Scene Graphs}}: {{Actionable Spatial Perception}} with {{Places}}, {{Objects}}, and {{Humans}},'' Jun. 2020.

\bibitem{greveCollaborativeDynamic3D2024}
E.~Greve, M.~B{\"u}chner, N.~V{\"o}disch, W.~Burgard, and A.~Valada, ``Collaborative {{Dynamic 3D Scene Graphs}} for {{Automated Driving}},'' in \emph{2024 {{IEEE International Conference}} on {{Robotics}} and {{Automation}} ({{ICRA}})}, May 2024, pp. 11\,118--11\,124.

\bibitem{ranaSayPlanGroundingLarge2023}
K.~Rana, J.~Haviland, S.~Garg, J.~{Abou-Chakra}, I.~Reid, and N.~Suenderhauf, ``{{SayPlan}}: {{Grounding Large Language Models}} using {{3D Scene Graphs}} for {{Scalable Robot Task Planning}},'' Sep. 2023.

\bibitem{waldRIO3DObject2019b}
J.~Wald, A.~Avetisyan, N.~Navab, F.~Tombari, and M.~Niessner, ``{{RIO}}: {{3D Object Instance Re-Localization}} in {{Changing Indoor Environments}},'' in \emph{2019 {{IEEE}}/{{CVF International Conference}} on {{Computer Vision}} ({{ICCV}})}.\hskip 1em plus 0.5em minus 0.4em\relax Seoul, Korea (South): IEEE, Oct. 2019, pp. 7657--7666.

\bibitem{zhuLivingScenesMultiobject2024}
L.~Zhu, S.~Huang, K.~Schindler, and I.~Armeni, ``Living {{Scenes}}: {{Multi-object Relocalization}} and {{Reconstruction}} in {{Changing 3D Environments}},'' in \emph{2024 {{IEEE}}/{{CVF Conference}} on {{Computer Vision}} and {{Pattern Recognition}} ({{CVPR}})}.\hskip 1em plus 0.5em minus 0.4em\relax Seattle, WA, USA: IEEE, Jun. 2024, pp. 28\,014--28\,024.

\bibitem{boreDetectionTrackingGeneral2019}
N.~Bore, J.~Ekekrantz, P.~Jensfelt, and J.~Folkesson, ``Detection and {{Tracking}} of {{General Movable Objects}} in {{Large Three-Dimensional Maps}},'' \emph{IEEE Transactions on Robotics}, vol.~35, no.~1, pp. 231--247, Feb. 2019.

\bibitem{schmidPanopticMultiTSDFsFlexible2022}
L.~Schmid, J.~Delmerico, J.~L. Schonberger, J.~Nieto, M.~Pollefeys, R.~Siegwart, and C.~Cadena, ``Panoptic {{Multi-TSDFs}}: A {{Flexible Representation}} for {{Online Multi-resolution Volumetric Mapping}} and {{Long-term Dynamic Scene Consistency}},'' in \emph{2022 {{International Conference}} on {{Robotics}} and {{Automation}} ({{ICRA}})}.\hskip 1em plus 0.5em minus 0.4em\relax Philadelphia, PA, USA: IEEE, May 2022, pp. 8018--8024.

\bibitem{qianPOCDProbabilisticObjectLevel2022a}
J.~Qian, V.~Chatrath, J.~Yang, J.~Servos, A.~P. Schoellig, and S.~L. Waslander, ``{{POCD}}: {{Probabilistic Object-Level Change Detection}} and {{Volumetric Mapping}} in {{Semi-Static Scenes}},'' in \emph{Robotics: {{Science}} and {{Systems XVIII}}}.\hskip 1em plus 0.5em minus 0.4em\relax {Robotics: Science and Systems Foundation}, Jun. 2022.

\bibitem{schmidKhronosUnifiedApproach2024b}
L.~Schmid, M.~Abate, Y.~Chang, and L.~Carlone, ``Khronos: {{A Unified Approach}} for {{Spatio-Temporal Metric-Semantic SLAM}} in {{Dynamic Environments}},'' in \emph{Robotics: {{Science}} and {{Systems XX}}}.\hskip 1em plus 0.5em minus 0.4em\relax {Robotics: Science and Systems Foundation}, Jul. 2024.

\bibitem{kohlbrecherFlexibleScalableSLAM2011}
S.~Kohlbrecher, O.~{von Stryk}, J.~Meyer, and U.~Klingauf, ``A flexible and scalable {{SLAM}} system with full {{3D}} motion estimation,'' in \emph{2011 {{IEEE International Symposium}} on {{Safety}}, {{Security}}, and {{Rescue Robotics}}}, Nov. 2011, pp. 155--160.

\bibitem{macenskiSLAMToolboxSLAM2021}
S.~Macenski and I.~Jambrecic, ``{{SLAM Toolbox}}: {{SLAM}} for the dynamic world,'' \emph{Journal of Open Source Software}, vol.~6, no.~61, p. 2783, May 2021.

\bibitem{huaiRobocentricVisualInertial2022}
Z.~Huai and G.~Huang, ``Robocentric visual--inertial odometry,'' \emph{The International Journal of Robotics Research}, vol.~41, no.~7, pp. 667--689, Jun. 2022.

\bibitem{crouseImplementing2DRectangular2016}
D.~F. Crouse, ``On implementing {{2D}} rectangular assignment algorithms,'' \emph{IEEE Transactions on Aerospace and Electronic Systems}, vol.~52, no.~4, pp. 1679--1696, Aug. 2016.

\bibitem{changGOATGOAny2023}
M.~Chang, T.~Gervet, M.~Khanna, S.~Yenamandra, D.~Shah, S.~Y. Min, K.~Shah, C.~Paxton, S.~Gupta, D.~Batra, R.~Mottaghi, J.~Malik, and D.~S. Chaplot, ``{{GOAT}}: {{GO}} to {{Any Thing}},'' Nov. 2023.

\bibitem{schroffFaceNetUnifiedEmbedding2015}
F.~Schroff, D.~Kalenichenko, and J.~Philbin, ``{{FaceNet}}: {{A}} unified embedding for face recognition and clustering,'' in \emph{2015 {{IEEE Conference}} on {{Computer Vision}} and {{Pattern Recognition}} ({{CVPR}})}.\hskip 1em plus 0.5em minus 0.4em\relax Boston, MA, USA: IEEE, Jun. 2015, pp. 815--823.

\bibitem{tanEfficientNetRethinkingModel2019}
M.~Tan and Q.~Le, ``{{EfficientNet}}: {{Rethinking Model Scaling}} for {{Convolutional Neural Networks}},'' in \emph{Proceedings of the 36th {{International Conference}} on {{Machine Learning}}}.\hskip 1em plus 0.5em minus 0.4em\relax PMLR, May 2019, pp. 6105--6114.

\bibitem{jonkerShortestAugmentingPath1988}
R.~Jonker and T.~Volgenant, ``A shortest augmenting path algorithm for dense and sparse linear assignment problems,'' in \emph{{{DGOR}}/{{NSOR}}}, H.~Schellhaas, P.~{van Beek}, H.~Isermann, R.~Schmidt, and M.~Zijlstra, Eds.\hskip 1em plus 0.5em minus 0.4em\relax Berlin, Heidelberg: Springer, 1988, pp. 622--622.

\bibitem{yolo11_ultralytics}
G.~Jocher and J.~Qiu, ``Ultralytics {{YOLO11}},'' 2024.

\bibitem{linMicrosoftCOCOCommon2014}
T.-Y. Lin, M.~Maire, S.~Belongie, J.~Hays, P.~Perona, D.~Ramanan, P.~Doll{\'a}r, and C.~L. Zitnick, ``Microsoft {{COCO}}: {{Common Objects}} in {{Context}},'' in \emph{Computer {{Vision}} -- {{ECCV}} 2014}, D.~Fleet, T.~Pajdla, B.~Schiele, and T.~Tuytelaars, Eds.\hskip 1em plus 0.5em minus 0.4em\relax Cham: Springer International Publishing, 2014, pp. 740--755.

\bibitem{sarlinSuperGlueLearningFeature2020}
P.-E. Sarlin, D.~DeTone, T.~Malisiewicz, and A.~Rabinovich, ``{{SuperGlue}}: {{Learning Feature Matching With Graph Neural Networks}},'' in \emph{Proceedings of the {{IEEE}}/{{CVF Conference}} on {{Computer Vision}} and {{Pattern Recognition}}}, 2020, pp. 4938--4947.

\end{thebibliography}

\end{document}